\pdfoutput=1
\documentclass[10pt,twocolumn,letterpaper]{article}
\usepackage{playcompass}
\usepackage{multirow}
\usepackage{pifont}
\usepackage{cuted}

\definecolor{linkblue}{rgb}{0.21,0.49,0.74}

\newcommand{\Nmain}{55\xspace}
\newcommand{\Nabl}{24\xspace}

\newcommand{\ours}{PlayCompass\xspace}

\newcommand{\arm}[1]{\textsf{#1}}
\newcommand{\native}{Native\xspace}
\newcommand{\nativeshow}{Native+Demos\xspace}
\newcommand{\fixedgated}{Fixed-Verify\xspace}
\newcommand{\evogated}{Evolving-Verify\xspace}
\newcommand{\noevo}{\arm{ours-no-evolution}\xspace}
\newcommand{\staticr}{\arm{ours-static-rubric}\xspace}
\newcommand{\nothemes}{\arm{ours-no-themes}\xspace}
\newcommand{\textfb}{\arm{ours-text-feedback}\xspace}

\newcommand{\Rz}{\mathcal{R}_0}
\newcommand{\Th}{\Theta}
\newcommand{\Ev}{E}
\newcommand{\cmark}{\ding{51}}
\newcommand{\xmark}{\ding{55}}

\usepackage[pagebackref,breaklinks,colorlinks,allcolors=linkblue,
  pdftitle={PlayCompass: Recursive Self-Improvement for Game Generation with Visually Grounded Evolving Rubrics},
  pdfauthor={Fuxiang Zhai, Sixiang Chen, Tengjun Huang, Lei Zhu}]{hyperref}

\title{PlayCompass: Recursive Self-Improvement for Game Generation with Visually Grounded Evolving Rubrics}

\author{Fuxiang Zhai$^{1}$\qquad Sixiang Chen$^{1}$\qquad Tengjun Huang$^{1}$\qquad Lei Zhu$^{1,2}$\\
$^{1}$The Hong Kong University of Science and Technology (Guangzhou)\\
$^{2}$The Hong Kong University of Science and Technology\\
{\tt\small fzhai598@connect.hkust-gz.edu.cn}\\
{\small Project page: \url{https://fuxiangzhai.github.io/PlayCompass/}}\vspace{-6pt}
}

\begin{document}
\maketitle
\begin{strip}
\centering
\includegraphics[width=\textwidth]{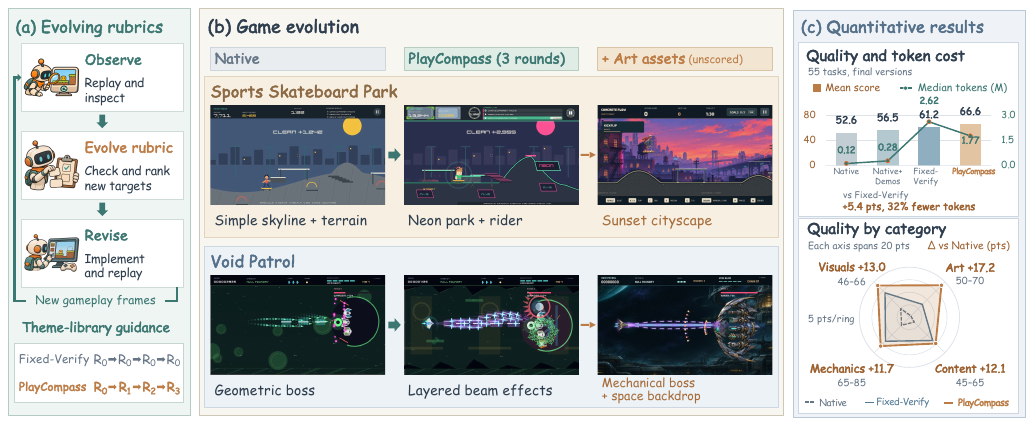}
\captionsetup{hypcap=false}
\captionof{figure}{\textbf{\ours improves games by evolving visually grounded rubrics.}
\textbf{(a)}~Rubrics evolve from replay frames and theme-library guidance to steer game revision.
\textbf{(b)}~Two Native games improve in just three \ours rounds (rightmost versions add art assets and remain unscored).
\textbf{(c)}~Across \Nmain tasks, \ours gains 14.0 points over Native and outperforms the fixed-rubric loop by 5.4 points with 32\% lower median billable token use (task means over two development runs for \ours and the fixed-rubric loop).}
\captionsetup{hypcap=true}
\label{fig:teaser}%
\end{strip}

\begin{abstract}
Coding agents can turn text briefs into playable games whose initial builds often lack content depth and visual polish. A fixed rubric can miss opportunities for improvement that emerge as a game changes, and short gameplay recordings can omit the states needed to verify an improvement.
We present \textbf{\ours}, a training-free framework for recursive game development that pairs a fixed anchor rubric for core requirements with an evolving rubric for improvement targets.
Each round, \ours uses gameplay frames from the current build to propose new targets and check which targets have been met. It ranks open targets using a theme library distilled from scored tasks outside the evaluation set and frozen before evaluation.
After each accepted revision, showcase demos capture the core loop and changes to inform the next rubric update.
On \Nmain GameCraft-Bench tasks with one model in every role, \ours improves the native coding agent on every task, by 14.0 points on average, and outperforms a fixed-rubric loop by 5.4 points with 32\% fewer tokens; human raters prefer its games on 25 of 30 tasks.
\end{abstract}

\section{Introduction}
\label{sec:intro}

A game is judged by what players see on screen, not by what its code contains.
Coding agents can generate playable games from natural-language briefs in real game engines~\cite{luo2026gamecraft,jiang2026opengame,chen2026swe,team2026game,dong2026code,wang2025openhands}.
Their builds can offer limited content and rely on placeholder graphics.
GameCraft-Bench~\cite{luo2026gamecraft} replays submitted input traces and scores the resulting gameplay recordings against hidden expert rubrics tailored to the intended experience of each game.
On our \Nmain tasks in our study, the initial Codex submissions average 69.5/100 for core mechanics, 47.7/100 for content depth, and 50.8/100 for art and presentation (\cref{tab:main}).

A self-improvement loop lets the agent play its current build, inspect the resulting gameplay, and use that evidence to guide code revisions~\cite{wu2026rsigame,chen2026recursive,yan2026harness,huang2026gui,jolicoeur2025multi,novikov2025alphaevolve}.
The loop needs its own criteria to decide what to change and check whether each revision meets its goal.
The game brief sets the core requirements. A rubric written solely from the brief can miss opportunities for improvement that emerge as the game evolves.
Verification also depends on which states a recording captures. A replay of the opening level offers little evidence about a mechanic that only appears in a later stage, even when it works correctly.
How can the loop use gameplay evidence to update and verify its improvement targets while preserving the brief's core requirements?

\begin{figure*}[t]
\centering
\includegraphics[width=0.96\textwidth]{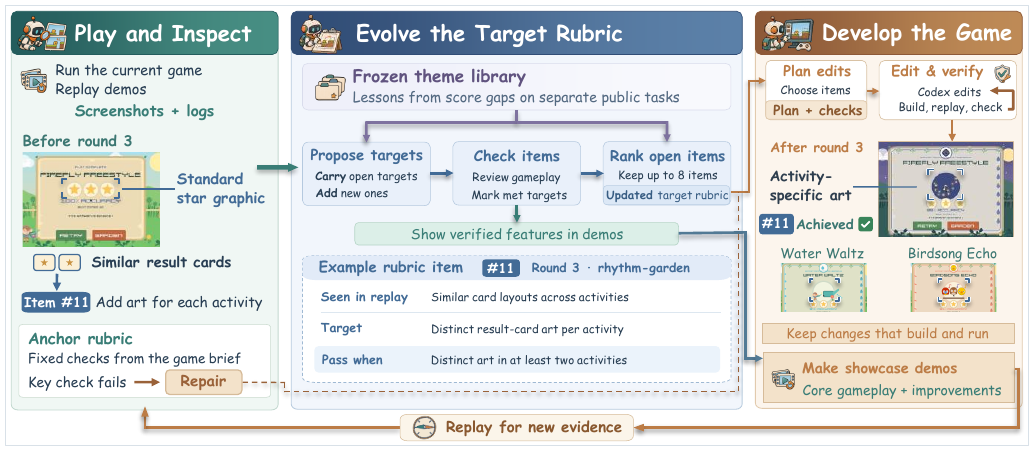}
\caption{\textbf{\ours evolves the game and its target rubric together.}
Gameplay frames guide target proposal and review.
A frozen theme library guides target proposal and ranking.
The designer turns selected criteria into a development plan and an acceptance contract that specifies what should appear on screen and how to reach it.
Showcase replays from the retained build provide fresh visual evidence for the next rubric update (bottom arrow).
In round~3 of \texttt{rhythm-garden}, similar result cards (left) motivate target~\#11 (center).
Water Waltz and Birdsong Echo replays (right) show distinct result-card illustrations in at least two activities, satisfying the criterion.}
\label{fig:pipeline}
\end{figure*}

We introduce \textbf{\ours}, a training-free framework for recursive self-improvement that uses visual feedback to evolve the game and its target rubric together while keeping the brief's core requirements in a fixed anchor rubric (\cref{fig:pipeline}).
The anchor guides repairs, and the loop uses gameplay frames from the current build to propose new targets and check which targets have been met.
In our analysis, gameplay recordings more often provided sufficient evidence for art and presentation targets than for content targets, so we focus new proposals on that category.
The loop prioritizes open targets using a frozen theme library distilled from public tasks outside the evaluation set.
Within each source task, we compare higher-scoring and lower-scoring game versions to extract reusable lessons.
The coding agent pursues the selected targets. For each newly retained build, a presenter creates showcase demos of the core gameplay loop and visible changes.
Replays of these demos provide fresh visual evidence for the next rubric update.

On \Nmain GameCraft-Bench~\cite{luo2026gamecraft} tasks, \ours improves official scores by 14.0 points on average over the initial Codex submissions.
We compare \ours with Fixed-Verify, a fixed-rubric refinement loop, using the same starting games and per-round coding budget over three development rounds.
In this comparison, \ours gains 5.4 points on average and reduces median billable token use per task by 32\% (\cref{tab:main}).
\ours leads on 40 of the \Nmain tasks, with the largest category gain in art and presentation ($+10.8$ points).
In additional evaluations, \ours scores higher on average than its ablated variants and the official implementations of RSIGame~\cite{wu2026rsigame} and RGC~\cite{chen2026recursive} on \Nabl GameCraft-Bench tasks, gains 5.2 points on average over Fixed-Verify on 12 SWE-Game~\cite{chen2026swe} tasks, and receives more preference votes than Fixed-Verify on 25 of 30 tasks in a blind video study.

\noindent Our main contributions are as follows:
\begin{itemize}
    \item We introduce \ours, a training-free framework for recursive game development that evolves the game and its target rubric together. A fixed anchor rubric guides checks and repairs for the brief's core requirements. Each round, the loop uses current gameplay frames to propose targets and verify which targets have been met.

    \item We construct a frozen theme library by comparing higher-scoring and lower-scoring versions of the same public tasks. We distill their score differences and judge rationales into reusable design lessons that guide target proposal and ranking on new games.

    \item We show that \ours outscores native on all \Nmain GameCraft-Bench tasks, with a mean gain of 14.0 points. Against Fixed-Verify, a fixed-rubric loop with matched starting builds and per-round coding budgets, \ours gains 5.4 points on average while reducing median billable token use per task.

    \item We find that per-round target updates and cross-task themes each improve scores in component ablations. \ours also outscores the official implementations of RSIGame and RGC in our shared three-round GameCraft-Bench setup and improves on Fixed-Verify on 12 SWE-Game~\cite{chen2026swe} tasks.
\end{itemize}

\section{Related Work}
\label{sec:related}

\subsection{Agentic game generation}
Benchmarks for agentic game development cover targeted edits to existing projects~\cite{chi2026gamedevbench} and the construction of games from textual specifications~\cite{zhang2026v,jiang2026opengame,chen2026swe,NEURIPS2023_91edff07,NEURIPS2023_1b44b878,gallotta2024large,valevski2025diffusion}. GameCraft-Bench~\cite{luo2026gamecraft} evaluates Godot projects through replayed gameplay.
For each task, this protocol lets us compare initial and refined builds under the same official rubric.

In iterative development, gameplay evidence also guides what the agent changes next.
Gameplay recordings and GUI playtests capture in-game behavior~\cite{jolicoeur2025multi,huang2026gui}, and runtime state injection lets verifiers test specific states directly~\cite{jia2026gamegen}.
A planner can use independent QA evidence to choose the next development increment~\cite{yan2026harness}.
RSIGame~\cite{wu2026rsigame} uses an evolving checklist of issues, improvement opportunities, priorities, and verification outcomes to choose the next development direction.
Recursive Game Creator (RGC)~\cite{chen2026recursive} uses gameplay trajectories and rendered observations to assess each build against game-specific criteria, recommend revisions, and select versions.
In \ours, a fixed anchor rubric guides checks and repairs for the brief's requirements.
Each round, the loop uses gameplay frames to propose new targets and check existing ones in a separate target rubric, then ranks open targets for subsequent revisions.

\subsection{Visual feedback and gameplay evidence}
Visual feedback can guide revisions to generated media and to the code that produces visual artifacts.
In image and video generation, feedback from candidate outputs guides prompt revisions or direct edits to the generated content~\cite{yang2024idea2img,wu2024self,long2026vista}.
For layouts, 3D scenes, and webpages produced by code, rendered outputs can guide targeted code revisions or provide a training signal~\cite{guo2026seeing,hu2024scenecraft,sansford2026vision,li2026relook,huang2024blenderalchemy,si2025design2code}.

In game development, visual observations from interactive tests help diagnose failures and guide repairs to generated builds~\cite{peng2026playcoder,team2026game}.
What a gameplay recording reveals depends on the actions taken during play.
A replay may reach the relevant state and still end before the specific player interaction needed to assess a target, leaving the reviewer with insufficient evidence.
To compare behavior across versions, \ours replays selected demos from the previous build with identical inputs.
For each newly retained build, a presenter also scripts new showcase demos that show the core gameplay loop and target-related changes, providing fresh evidence for the next round of target checks.

\subsection{Evolving rubrics and target prioritization}
Rubrics specify explicit criteria for evaluating open-ended outputs~\cite{chen2026holistic}.
Text-to-image evaluators translate prompt requirements into questions about objects, attributes, and relations~\cite{hu2023tifa,cho2024davidsonian,huang2024vbench}.
VideoArgus~\cite{zeng2026videoargus} constructs a rubric for each input and applies it to all corresponding candidate videos to enable consistent comparisons.
Criterion-level judgments can also be combined into rewards for reinforcement learning in language and image generation~\cite{gunjal2026rubrics,feng2026rubricrl}.

Eureka~\cite{ma2024eureka} uses feedback from policy training to revise reward programs, and evolving-rubric methods adapt evaluation criteria during language model policy optimization~\cite{guan2026evorubric,ding2026evorubrics,wang2026serpo,pmlr-v267-wang25bx,romera2024mathematical}.
RuHarness~\cite{chen2026critic} updates behavioral criteria from execution trajectories and prioritizes unresolved weaknesses to guide harness revisions while keeping the task objective fixed.
\ours uses gameplay evidence to update a separate rubric of visually verifiable targets for the next game revision.
A fixed anchor rubric guides checks and repairs for the brief's requirements.

When several targets remain open, experience from other tasks can help decide which to pursue first.
Voyager~\cite{wang2023voyager} reuses executable skills, and ExpeL~\cite{zhao2024expel} distills transferable lessons from past trajectories, including comparisons of successful and failed attempts on the same task.
\ours compares higher-scoring and lower-scoring game versions within public tasks outside the evaluation set and distills their score differences and judge rationales into a frozen theme library.
The loop combines these theme priors with category weights and each target's current review verdict to rank open targets for subsequent revisions.

\section{Method}
\label{sec:method}

\noindent{\textbf{Task and official score.}}
Each GameCraft-Bench~\cite{luo2026gamecraft} task provides a text brief $b$, a Godot workspace, and shared 2D assets.
A submission consists of a game project $g$ and a set $\mathcal{D}$ of up to ten gameplay input traces (demos).
Let $\mathrm{BUILD}(g,\mathcal{D})\in\{0,1\}$ indicate whether $g$ launches and at least one demo parses.
The verifier replays each valid trace $\tau$ in a fresh Godot instance, recording up to 20\,s as $\rho(g,\tau)$, and a multimodal judge scores the recordings against a hidden expert rubric, yielding category scores $s_k(g,\mathcal{D})\in[0,1]$ and the official score
\begin{equation}
S(g,\mathcal{D}) =
\begin{cases}
\sum_k w_k\,s_k(g,\mathcal{D}), & \mathrm{BUILD}(g,\mathcal{D})=1,\\
0, & \text{otherwise}.
\end{cases}
\label{eq:official}
\end{equation}
The public weights $w=(0.15,0.35,0.15,0.35)$ correspond to core mechanics, content depth, functional visuals, and art and presentation.
The judge averages persistent properties over demos but takes the maximum for scenario-specific requirements~\cite{luo2026gamecraft}, which thus earn credit only if some demo reaches the relevant state.
Refinement never accesses the hidden rubric or official scores.

\noindent{\textbf{Overview.}}
Starting from a Codex submission $(g_0,\mathcal{D}_0)$, \ours refines the game and its demos over $R$ rounds without training (\cref{fig:pipeline}).
It keeps a fixed anchor rubric $\Rz$ for the brief's requirements, open targets $\mathcal{T}_r$, fulfilled targets $\mathcal{F}_r$, and a frozen theme library $\Th$.
With $\mathcal{T}_0=\mathcal{F}_0=\varnothing$ and round-zero evidence $\Ev_0$, round $r$ performs
\begin{align}
(\mathcal{T}_r,\mathcal{F}_r)
    &= \textsc{Evolve}(\mathcal{T}_{r-1},\mathcal{F}_{r-1},
        \Ev_{r-1};\Rz,\Th), \nonumber\\
\tilde g_r
    &= \textsc{Develop}(g_{r-1},\Rz,\mathcal{T}_r,\Ev_{r-1}), \nonumber\\
g_r &= \textsc{Keep}(g_{r-1},\tilde g_r), \label{eq:loop}\\
(\mathcal{D}_r,\Ev_r) &= \textsc{Observe}(g_r;\mathcal{T}_r,\mathcal{F}_r). \nonumber
\end{align}
If \textsc{Keep} retains $g_{r-1}$, we reuse $\mathcal{D}_{r-1}$ and $\Ev_{r-1}$, and we finally submit $(g_R,\mathcal{D}_R)$.
All roles share one model: a target proposer and a reviewer maintain the rubric, a designer plans rounds, Codex edits the game, a QA checker tests revisions, and a presenter writes showcase demos.

\subsection{Gameplay evidence for refinement}
\label{sec:visual}

For each newly retained build $g_r$, \textsc{Observe} collects sandbox frames and runtime logs $\sigma(g_r,\tau)$ and benchmark replay recordings $\rho(g_r,\tau)$ of traces $\tau$:
\begin{equation}
\begin{aligned}
\Ev_r={}&\big\{\sigma(g_r,\tau):\tau\in P\cup\Pi_r\big\}\\
&\cup\big\{\rho(g_r,\tau):\tau\in\mathcal{D}_r\cup\bar{\mathcal{D}}_{r-1}\big\}.
\end{aligned}
\label{eq:observe}
\end{equation}
The four probes $P$ are traces scripted from $b$ in round zero and replayed on every build.
A player policy, also written in round zero, reads game-reported state to produce six episodes $\Pi_r$ without per-action model calls; this state never counts as evidence.
The demo set $\mathcal{D}_r$ holds up to five showcase demos for $g_r$, followed by inherited Codex demos that still replay, and $\bar{\mathcal{D}}_{r-1}$ replays the previous build's first three showcase demos to compare builds under identical inputs.

The presenter scripts showcase demos from the brief, source code, screenshots, and target highlights: the first shows the core loop, and later ones show up to six implemented highlights in normal play.
The four consecutive frames with the largest pixel change in each showcase recording form a \emph{motion strip}, and probes fill $\mathcal{D}_R$ up to the ten-trace limit for the final submission.

An internal vision-language model (VLM) reviewer checks each criterion $c$ using only frames and logs from $\Ev$ (at most 24 of about 700 frames per call).
It returns a verdict $y(c;\Ev)\in\{\textsf{pass},\textsf{fail},\varnothing\}$.
A \textsf{pass} requires observed evidence that the game satisfies $c$, a \textsf{fail} requires an observed violation, and $\varnothing$ denotes insufficient evidence for either.
Each \textsf{pass} or \textsf{fail} cites supporting frames or log lines.
\ours uses this \emph{lean} review by default; in \emph{full review}, used by the verified baselines (\cref{sec:setup}) and an eight-task pilot of \ours, an evidence agent also writes up to three traces per round to reach states of criteria marked $\varnothing$, so that the reviewer can re-check them.

\subsection{Evolving the target rubric}
\label{sec:rubric}

\noindent{\textbf{Anchor rubric $\Rz$.}}
In round zero, we derive $\Rz$ (at most 16 criteria) from $b$, assign each criterion a public category, mark those the brief makes mandatory, and check $g_0$ using $\Ev_0$.
$\Rz$ stays fixed in \ours and the fixed-rubric baseline, and its core-mechanics and mandatory criteria form $\Rz^{\mathrm{core}}$, any \textsf{fail} of which triggers repair (\cref{eq:direction}).

\noindent{\textbf{Theme library $\Th$ (across tasks).}}
We build $\Th$ offline from public tasks disjoint from the evaluation set, pairing officially scored versions $p=(g^+,g^-)$ of the same task whose scores differ by at least 5 points (90 pairs from 23 tasks).
For each pair, we retain up to three requirements with the largest positive gaps from $g^-$ to $g^+$, weighted by their contribution to $S$, with the judge's rationales for both versions.
A meta agent distills these comparisons into 23 reusable, genre-tagged themes grouped by category, each traceable to its supporting pairs.

For pair $p$, let $\Delta_k(p)$ be the sum of its retained weighted gaps in category $k$.
For a theme $h$ in category $k$, let $\mathcal{M}_h$ contain its supporting pairs and $\mathcal{M}_k$ all pairs with a retained requirement in $k$.
The theme prior is
\begin{equation}
\theta(h)=\operatorname{clip}_{[0.5,\,1.5]}\left(\frac{\operatorname{mean}_{p\in\mathcal{M}_h}\Delta_k(p)}{\operatorname{mean}_{p\in\mathcal{M}_k}\Delta_k(p)}\right).
\label{eq:theta}
\end{equation}
A prior above $1$ means that the supporting pairs' mean gap exceeds the category baseline, and clipping limits the influence of themes supported by few pairs.
We validate and freeze the library before evaluation (\cref{sec:analysis}).

\noindent{\textbf{Target rubric (within a task).}}
A \emph{target} specifies an improvement verifiable in normal play.
Before round $r$, the target proposer suggests up to four targets $\mathcal{N}_r$ from the frames of $g_{r-1}$ in $\Ev_{r-1}$, with showcase frames first.
It applies genre-matched themes that fit $b$ and address gaps visible in the frames.
Each proposal specifies a visible outcome, where to verify it, its source (frames, brief, or theme), and an optional theme $h_t$; debug overlays, test-only hooks, and content removal are excluded.
\ours proposes targets only for art and presentation, whose changes recordings verify far more often than content changes (\cref{sec:analysis}); other categories rely on anchor criteria.

$\textsc{Dedup}_\delta$ drops proposals whose word-level Jaccard similarity to an anchor criterion, an earlier target, or an earlier retained proposal is at least $\delta=0.5$, and the reviewer checks the rest and the open targets against $\Ev_{r-1}$:
\begin{align}
\mathcal{C}_r &= \mathcal{T}_{r-1}\cup\textsc{Dedup}_\delta\left(\mathcal{N}_r;\Rz\cup\mathcal{T}_{r-1}\cup\mathcal{F}_{r-1}\right),\nonumber\\
\mathcal{F}_r &= \mathcal{F}_{r-1}\cup\big\{t\in\mathcal{C}_r:y(t;\Ev_{r-1})=\textsf{pass}\big\},\label{eq:evolve}\\
\mathcal{T}_r &= \operatorname{top}_K\nolimits^{\psi}\big(\mathcal{C}_r\setminus\mathcal{F}_r\big).\nonumber
\end{align}
The open set keeps up to $K=8$ targets marked \textsf{fail} or $\varnothing$, ranked by
\begin{equation}
\psi(t)=\theta(h_t)\,w_{k_t}\,\gamma(t),\qquad
\gamma(t)=\begin{cases}1, & y(t;\Ev_{r-1})=\textsf{fail},\\0.7, & y(t;\Ev_{r-1})=\varnothing,\end{cases}
\label{eq:priority}
\end{equation}
where $k_t$ is the category of $t$ and $\theta(h_t)=1$ for untagged targets.
The designer sees open targets in this order; $w_{k_t}$ is constant for \ours, which proposes only art targets, but matters for variants with content targets (\cref{sec:variants}).
Targets beyond the top $K$ are dropped but may be proposed again, and fulfilled targets are not re-checked; instead, they supply highlights for later showcase demos.

\subsection{Recursive game development}
\label{sec:rsi}

\noindent{\textbf{Direction.}}
Each round repairs the build or extends one category $k\in\{\textsf{art},\textsf{content}\}$, where $n_k(r)$ counts earlier extensions of $k$.
The preferred direction is
\begin{equation}
\hat{d}_r=\begin{cases}
\textsf{repair}, & \exists\,c\in\Rz^{\mathrm{core}}:y(c;\Ev_{r-1})=\textsf{fail},\\
\textsf{art}, & \text{else if }n_{\textsf{art}}(r)\leq n_{\textsf{content}}(r),\\
\textsf{content}, & \text{otherwise}.
\end{cases}
\label{eq:direction}
\end{equation}
Repairs take precedence, and extensions alternate, starting with art.
A category is eligible if it has an open target or an anchor criterion marked \textsf{fail} or $\varnothing$.
The round direction $d_r$ falls back to the other category if $\hat{d}_r$ is ineligible, and to deepening passed content criteria (or passed art criteria if none exist) if neither category is eligible.

\noindent{\textbf{Develop.}}
Given $d_r$, the designer selects criteria $\mathcal{R}^{\star}_r\subseteq\Rz\cup\mathcal{T}_r$ (anchor failures for a repair; eligible anchor criteria and top-ranked targets of the category for an extension) and writes an \emph{acceptance contract} of three to eight checks, each naming an onscreen outcome and the inputs that reach it.
Codex edits a sandboxed copy of $g_{r-1}$ and writes a QA trace per check in up to three sessions sharing a 20-minute budget.
After each session, the QA checker builds the game, replays the traces, and marks each check \emph{done}, \emph{not done}, or \emph{not visible}.
Editing stops when all checks are done, a session changes nothing or yields no report, or a limit is reached, and the last candidate is $\tilde g_r$.

\noindent{\textbf{Version retention.}}
\textsc{Keep} sets $g_r=\tilde g_r$ if the candidate passes build and runtime checks and differs from $g_{r-1}$ (project hash), and keeps $g_{r-1}$ otherwise.
\ours thus retains valid changes that the reviewer cannot yet confirm, which verified acceptance would reject at a cost in official score (\cref{sec:analysis}).
Verified baselines accept a candidate only after rubric checks and blind pairwise comparisons with the incumbent in both presentation orders.

\section{Experiments}
\label{sec:experiments}

\subsection{Setup}
\label{sec:setup}
\begin{table*}[t]
\centering
\footnotesize
\setlength{\tabcolsep}{3.0pt}
\newcommand{\up}[1]{\,{\scriptsize\textcolor{green!45!black}{#1}}}
\caption{\textbf{Main results on \Nmain GameCraft-Bench tasks.} Official scores (points) are task means, with category weights in parentheses. Green: gain of \ours over \fixedgated. $\Delta$: mean paired difference in the overall score, with a 95\% bootstrap CI over tasks for the comparison with \fixedgated. Tok.: median billable tokens per task. h: median wall-clock hours per run. All methods use GPT-5.6-sol and start from the \native game.}
\label{tab:main}
\begin{tabular}{@{}lccccccccc@{}}
\toprule
& & Core mech. & Content depth & Func.\ visuals & Art \& pres. & \multicolumn{2}{c}{Overall $\Delta$} & & \\
\cmidrule(lr){7-8}
Method & Overall & (M, 0.15) & (D, 0.35) & (V, 0.15) & (A, 0.35) & vs.\ \native & vs.\ \fixedgated\ [95\% CI] & Tok. & h \\
\midrule
\native (Codex)                    & 52.6 & 69.5 & 47.7 & 50.9 & 50.8 & -- & -- & 0.12M & -- \\
\nativeshow (our demos)            & 56.5 & 76.5 & 52.4 & 56.3 & 52.1 & $+4.0$ & -- & 0.28M & 0.7 \\
\fixedgated (fixed rubric)         & 61.2 & 78.5 & 58.1 & 59.8 & 57.3 & $+8.6$ & -- & 2.62M & 2.5 \\
\ours (ours)                       & \textbf{66.6}\up{+5.4} & \textbf{81.2}\up{+2.7} & \textbf{59.8}\up{+1.7} & \textbf{63.9}\up{+4.1} & \textbf{68.1}\up{+10.8} & $\mathbf{+14.0}$ & $\mathbf{+5.4}$ {\scriptsize[3.1, 7.9]} & 1.77M & 1.7 \\
\bottomrule
\end{tabular}
\end{table*}

\noindent{\textbf{Tasks.}}
We split the 140 GameCraft-Bench~\cite{luo2026gamecraft} tasks into disjoint pools for main evaluation, development (six tasks), and theme-library construction and validation.
The \Nmain main tasks span all 15 genres and the native-score tertiles (20 low, 18 middle, 17 high); a \Nabl-task ablation subset is drawn by a fixed hash within each tertile.
We configured \ours after the verified loops had run on 29 of these tasks (\cref{sec:analysis}) and also report the other 26.

\noindent{\textbf{Baselines.}}
\textbf{\native} is the initial Codex submission $(g_0,\mathcal{D}_0)$ from the official harness.
\textbf{\nativeshow} keeps $g_0$ but submits presenter-authored showcase demos first, isolating the effect of demos.
\textbf{\fixedgated}, our primary iterative baseline, uses the same evidence with the fixed anchor rubric $\Rz$.
It repairs on core \textsf{fail}s and otherwise extends whichever of content depth and art and presentation has the larger weighted gap in $\Rz$.
Following RGC~\cite{chen2026recursive}, it accepts a candidate only after rubric checks and blind pairwise comparisons with the incumbent in both presentation orders, and it uses full review.
Both loops share $g_0$, the frozen round-zero package (anchor rubric, probes, player policy, and initial evidence), the presenter, and the asset library, and run three rounds with a 20-minute Codex budget each; \cref{sec:variants} separates their differences in rubric, scheduling, and acceptance.
On the ablation subset, we also run the official RSIGame~\cite{wu2026rsigame} and RGC code once per task from $g_0$ for three rounds and score each export with its own demos.

\noindent{\textbf{Evaluation.}}
The official verifier records demo replays in a network-isolated sandbox, and the unmodified upstream scorer applies the hidden rubric with GPT-5.6.
All agents use GPT-5.6-sol without image or 3D generation.
We run \ours and \fixedgated twice per task (\evogated twice on the 29 design tasks; other methods once); repeated runs share the native game and round-zero package but make fresh model calls.
Each export is scored twice on the $100\,S$ scale (\cref{eq:official}), using two independent recordings for real-time tasks and two passes over one recording otherwise.
We report paired task-level differences with 95\% bootstrap CIs (10{,}000 resamples), two-sided Wilcoxon tests, and win/tie/loss (W/T/L) counts.

\subsection{Main results}
\label{sec:results}

\begin{table}[t]
\centering
\footnotesize
\setlength{\tabcolsep}{3pt}
\caption{\textbf{Robustness of the gain} of \ours over \fixedgated (score, points). $n$: tasks. W/T/L: tasks on which \ours is better, tied, or worse. Bold: 95\% CI excludes 0.}
\label{tab:robust}
\begin{tabular}{@{}llccc@{}}
\toprule
& Setting & $n$ & $\Delta$ [95\% CI] & W/T/L \\
\midrule
& Main comparison (\cref{tab:main}) & \Nmain & \textbf{+5.4} [3.1, 7.9] & 40/0/15 \\
& Outside the design tasks & 26 & \textbf{+6.0} [2.8, 9.7] & 19/0/7 \\
\midrule
\multirow{3}{*}{Tertile} & low native score & 20 & \textbf{+9.3} [4.7, 14.6] & 16/0/4 \\
& mid native score & 18 & +3.2 [$-$0.3, 6.8] & 10/0/8 \\
& high native score & 17 & \textbf{+3.0} [1.4, 4.7] & 14/0/3 \\
\midrule
\multirow{2}{*}{Weights} & equal category weights & \Nmain & \textbf{+4.8} [2.6, 7.4] & 40/0/15 \\
& without art and pres. & \Nmain & \textbf{+2.5} [0.2, 5.0] & 34/0/21 \\
\midrule
\multirow{2}{*}{Judge} & Gemini-2.5-Flash & \Nmain & \textbf{+3.3} [0.9, 5.8] & 36/1/18 \\
& Gemini-3-Flash & \Nmain & +2.3 [$-$0.02, 4.7] & 33/0/22 \\
\midrule
Humans & blind preference & 30 & -- & 25/3/2 \\
Benchmark & SWE-Game & 12 & \textbf{+5.2} [0.2, 10.2] & 9/0/3 \\
\bottomrule
\end{tabular}
\end{table}

\begin{figure}[t]
\centering
\includegraphics{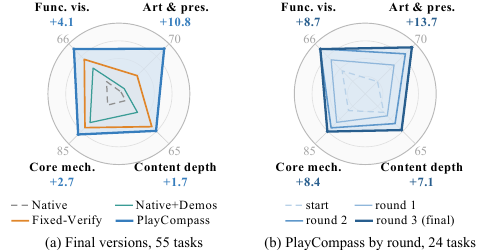}
\caption{\textbf{Category profiles} (task means, points; each axis spans 20 points, with the rim value in gray). (a)~Final versions on the \Nmain main tasks; numbers give \ours$-$\fixedgated. (b)~\ours after each round on the \Nabl ablation tasks; numbers give final$-$start.}
\label{fig:radar}
\end{figure}

\noindent{\textbf{\ours improves every game and outperforms the fixed-rubric loop.}}
Over three rounds, \ours raises the mean official score from 52.6 to 66.6 and improves on all \Nmain tasks (\cref{tab:main}).
Against \fixedgated, it achieves a mean paired gain of 5.4 points (95\% CI [3.1, 7.9]) and scores higher on 40 of the \Nmain tasks and in 12 of the 15 genres.
The gain concentrates in art and presentation ($+10.8$ points, \cref{fig:radar}a), where \ours proposes targets, while the content-depth gain ($+1.7$, 95\% CI [$-0.9$, 4.3]) is not significant (\cref{sec:analysis}).

\noindent{\textbf{Development adds to the demo gain.}}
Because the judge takes the maximum over demos for scenario-specific requirements, demos alone change the score: with the game unchanged, \nativeshow gains 4.0 points over \native (95\% CI [2.8, 5.3]).
Relative to \nativeshow, development adds 4.6 points for \fixedgated and 10.0 points for \ours (95\% CIs [3.2, 6.2] and [7.8, 12.4]).
These gains reflect both games and demos; \cref{sec:qual} compares the games under identical inputs.

\begin{figure*}[t]
\centering
\includegraphics{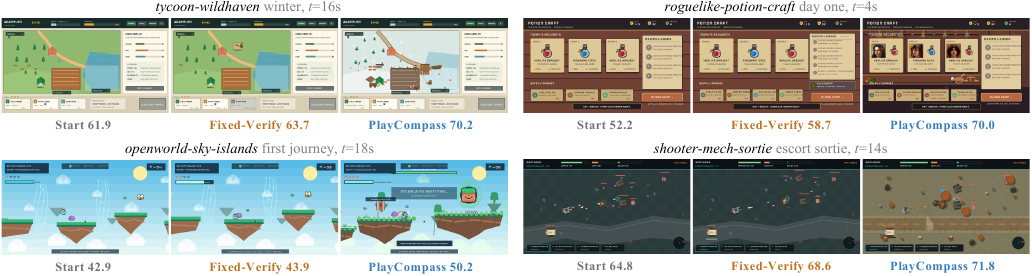}
\caption{\textbf{Final versions under identical inputs.} For each task, one showcase demo of the shared start (the native game with the presenter's demos) is replayed on the final versions of \fixedgated and \ours and cut at the same time; numbers are official scores. On the same screens, \ours adds a winter scene with a dock (\texttt{tycoon-wildhaven}), customer portraits (\texttt{roguelike-potion-craft}), planted islands with their guardian (\texttt{openworld-sky-islands}), and desert terrain (\texttt{shooter-mech-sortie}). \Cref{fig:teaser}b shows the rounds of \ours.}
\label{fig:qual}
\end{figure*}

\noindent{\textbf{The gain holds across tasks, weights, and judges.}}
The gain is positive in every native-score tertile and largest in the lowest ($+9.3$ points, \cref{tab:robust}), where initial builds leave the most room for improvement.
On the 26 tasks not used to configure \ours, it is 6.0 points [2.8, 9.7], and excluding art and presentation still leaves 2.5 points [0.2, 5.0].
Judges from another model family score all versions higher, compressing differences: the gain shrinks to 3.3 points [0.9, 5.8] with Gemini-2.5-Flash and 2.3 points [$-0.02$, 4.7] with Gemini-3-Flash, but both keep a significant art gain.
On 12 SWE-Game tasks~\cite{chen2026swe}, which weight required functionality at 85\%, the gain is 5.2 points [0.2, 10.2] (Wilcoxon $p=0.09$).

\noindent{\textbf{\ours uses 32\% fewer tokens.}}
Per task, \ours uses a median of 1.77M billable tokens and 1.7 hours, versus 2.62M and 2.5 hours for \fixedgated (\cref{tab:main}).
The savings come from verification rather than the rubric: reviews, pairwise comparisons, and evidence collection cost 0.48M versus 1.39M tokens per run, because \ours keeps every valid build and uses lean review, while Codex editing costs are comparable.

\noindent{\textbf{\ours outperforms RSIGame and RGC in our setting.}}
On the ablation subset (\cref{tab:variants}), \ours exceeds RSIGame by 11.7 points [8.6, 15.1] and RGC by 12.5 points [9.4, 15.7], scoring higher on 23 and all \Nabl tasks, with gains in all four categories (12.2 and 11.8 points in art and presentation).
Both run their official code with our model, start, budget, and judge, so the comparison reflects our setting, not their original setups.
Under their own rules, RSIGame keeps its starting version on 16 of the 24 tasks, and RGC gains only 2.3 points [$-0.9$, 5.8].

\subsection{Qualitative results and user study}
\label{sec:qual}

\noindent{\textbf{Final builds differ visibly under identical inputs.}}
Because \cref{fig:qual} replays the same showcase demo of the shared start on both final builds and cuts at the same time, the added scene art and illustrated interface elements come from the games rather than from the demos written for each version.

\noindent{\textbf{Human raters prefer \ours.}}
In a blind side-by-side study, 24 volunteer raters compared showcase videos of the first-run final builds of \ours and \fixedgated on 30 tasks, two per genre, drawn by a fixed hash before any video was watched.
Each task received four judgments in balanced left--right order, and raters chose a side or no clear difference.
By task-level majority, raters preferred \ours on 25 tasks and \fixedgated on two, with three ties (sign test, $p=5.6\times10^{-6}$); over 120 judgments, \ours won 88 to 9 with 23 ties (Krippendorff's $\alpha=0.78$).
On five-point scales, \ours was also rated higher for visual clarity ($+0.42$) and playability ($+0.24$), with slightly fewer bugs reported ($-0.12$).
Since each side shows its own showcase demos, these preferences concern the games as presented to players, rather than under identical inputs as in \cref{fig:qual}.

\subsection{Ablations}
\label{sec:variants}

\begin{table}[t]
\centering
\footnotesize
\setlength{\tabcolsep}{1.4pt}
\caption{\textbf{Ablations} on the \Nabl ablation tasks: variant minus \ours (points; 95\% CI for the overall score), where \ours scores 66.4. Frames: agents see recorded frames (\cmark) or only logs, state, and source (\xmark). Rubric: anchor only (fixed), evolving targets with or without themes, or targets proposed once (static). Keep: every valid build (all) or verified builds only (verif.). Middle: components of \ours. Bottom: other loops, including the official RSIGame and RGC code (one run each); \evogated, \fixedgated, RSIGame, and RGC are compared on main-experiment scores, where \ours scores 67.0. Bold: the CI excludes 0.}
\label{tab:variants}
\begin{tabular}{@{}lccc|cc@{}}
\toprule
Arm & Frames & Rubric & Keep & $\Delta$ overall & $\Delta$ art \\
\midrule
\ours & \cmark & $\mathcal{T}_r{+}\Th$ & all & 0 & 0 \\
\midrule
\textfb & \xmark & $\mathcal{T}_r{+}\Th$ & all & $\mathbf{-6.1}$ {\scriptsize[$-8.6$, $-3.6$]} & $\mathbf{-6.0}$ \\
\noevo & \cmark & fixed & all & $\mathbf{-4.7}$ {\scriptsize[$-8.0$, $-1.5$]} & $\mathbf{-6.8}$ \\
\staticr & \cmark & static & all & $\mathbf{-4.8}$ {\scriptsize[$-7.7$, $-2.0$]} & $\mathbf{-7.1}$ \\
\nothemes & \cmark & $\mathcal{T}_r$ & all & $\mathbf{-4.6}$ {\scriptsize[$-7.1$, $-2.3$]} & $\mathbf{-6.8}$ \\
\midrule
\evogated & \cmark & $\mathcal{T}_r{+}\Th$ & verif. & $\mathbf{-6.4}$ {\scriptsize[$-10.2$, $-3.0$]} & $\mathbf{-10.8}$ \\
\fixedgated & \cmark & fixed & verif. & $\mathbf{-7.8}$ {\scriptsize[$-12.1$, $-4.0$]} & $\mathbf{-12.3}$ \\
RSIGame & \multicolumn{3}{c|}{official code} & $\mathbf{-11.7}$ {\scriptsize[$-15.1$, $-8.6$]} & $-12.2$ \\
RGC & \multicolumn{3}{c|}{official code} & $\mathbf{-12.5}$ {\scriptsize[$-15.7$, $-9.4$]} & $-11.8$ \\
\nativeshow & \multicolumn{3}{c|}{no development} & $\mathbf{-9.8}$ {\scriptsize[$-12.8$, $-7.2$]} & $\mathbf{-13.7}$ \\
\bottomrule
\end{tabular}
\end{table}

All variants in \cref{tab:variants} share the native games and budget of \ours.
\textfb hides frames from development agents but keeps the presenter; \noevo uses only the anchor rubric; \staticr proposes targets once; \nothemes drops the theme library; and \evogated adds content and art targets to \fixedgated, keeping its scheduling and acceptance.

\noindent{\textbf{Visual feedback improves all four categories.}}
Without frames, the score drops by 6.1 points [3.6, 8.6] and \ours wins on 20 of \Nabl tasks.
All four categories fall, by 4.1 (M), 6.7 (D), 6.6 (V), and 6.0 (A) points, suggesting that frames help the designer and QA checker beyond targets.

\noindent{\textbf{Targets must be re-grounded on each build.}}
Removing targets, proposing them once, or dropping themes lowers the score by 4.6--4.8 points (art: 6.8--7.1).
One-shot targets are no better than none, so the gain comes from re-grounding targets on the frames of each new build.
Since targets also guide the presenter, these gaps reflect changes to both the games and their demos.

\noindent{\textbf{Retention and target evolution both matter.}}
On these tasks, \ours leads \evogated by 6.4 points (95\% CI [3.0, 10.2]) and \fixedgated by 7.8 points (95\% CI [4.0, 12.1]).
Evolving targets recover only 1.4 of these points in the verified loop but are worth 4.7 points in \ours, so they pay off when the loop keeps the builds that pursue them.
Internal reviews marked 65.9\% of selected art targets as achieved in 110 \ours runs, but only 25\% in 58 \evogated runs, which achieve 1.0 targets per run on average versus 4.0 for \ours.

\subsection{Analysis}
\label{sec:analysis}

\begin{figure}[t]
\centering
\includegraphics{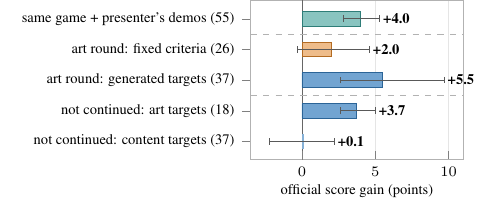}
\caption{\textbf{What the loop sees and targets determines what it gains.} Mean official score gains (points, 95\% CIs; counts in parentheses). Top: the unchanged native game with the presenter's demos (tasks). Bottom: single rounds relative to the build they extend, grouped by what they pursued (rounds): art rounds of \fixedgated (fixed criteria) and \evogated (generated targets), then rejected (not continued) \evogated candidates.}
\label{fig:gaps}
\end{figure}

We examine why the gains follow the evidence that recordings provide.
For each saved candidate of the verified loops, including rejected ones, the round gain is its official score minus that of the build entering the round, which \cref{fig:gaps} contrasts with the demo-only gain of \nativeshow.

\phantomsection
\noindent{\textbf{Gains accumulate over rounds.}}
\label{sec:compound}
On the ablation tasks, the mean score of \ours rises from 56.6 at round zero (\nativeshow) to 60.7, 64.4, and 66.4 after rounds 1--3 (one run per task, single-pass scoring, \cref{fig:radar}b).
Every final build beats its round-zero version, by 9.8 points on average [7.2, 12.8], led by art ($+13.7$) and content depth ($+7.1$).
Per-round gains shrink from 4.1 to 3.7 and 2.0 points but stay positive, so further rounds may still help.

\phantomsection
\noindent{\textbf{Recordings verify art targets far more often than content targets.}}
\label{sec:steering}
Across 58 \evogated runs on 29 tasks, the reviewer marked 18\% of art-target checks and 62\% of content-target checks as \emph{insufficient evidence}.
More specific targets widen rather than close the gap: in a six-task development comparison, they raised these rates from 15\% and 52\% to 31\% and 76\%.
Content often lives in states that a 20-second recording of a fresh game rarely reaches, whereas art is visible on almost every screen, which motivates proposing only art targets while content relies on anchor criteria.

\phantomsection
\noindent{\textbf{Verified acceptance discards useful art changes.}}
\label{sec:rubric-adds}
Rejected \evogated candidates still gained 3.7 official points on average (95\% CI [2.6, 5.0]) across 18 art-target rounds, but only 0.1 points [$-2.2$, 2.2] across 37 content-target rounds (\cref{fig:gaps}).
Gating on the reviewer thus rejects real art improvements, so \ours keeps every valid build.
Where \ours trails \fixedgated most, it still leads in art but misses content states, such as a second arena or a landed catch, that no 20-second demo reaches.

\noindent{\textbf{Gains are not fitted to the internal reviewer.}}
The final builds of \ours have a 9.5-point lower internal art pass rate than those of \fixedgated, since lean review runs without re-checks, yet they score higher on the official art requirements.
No arm uses an image generator, so the added art is drawn through the engine or taken from the asset library that all arms share.

\noindent{\textbf{Themes transfer lessons, not task details.}}
On six held-out public tasks, a theme matched the better version's advantage in 19 of 24 pairs, but in one pair after swapping the judge's rationales.
In \ours, 39\% of theme-derived proposals were achieved, versus 24\% of those from frames or the brief, and targets are less similar to the hidden art requirements than the brief-derived anchor rubric (TF-IDF 0.08 vs.\ 0.14), indicating no task-level leakage.

\section{Conclusion}
\label{sec:conclusion}

We presented \ours, a training-free framework that refines a game together with its target rubric: a fixed anchor rubric preserves the brief's requirements, while targets proposed and verified on each build's gameplay frames and ranked by a frozen theme library decide what to improve next.
On \Nmain GameCraft-Bench tasks, \ours improves every native game, gains 5.4 points on average over a fixed-rubric loop with 32\% fewer billable tokens, and is preferred by human raters on 25 of 30 tasks.
Our analysis suggests that self-improving loops gain most where their evidence can show success, and that keeping only revisions an internal reviewer can confirm discards real gains.

\noindent{\textbf{Limitations.}}
We study 2D Godot games over three rounds with one agent backbone, and the official judge shares its model family.
Content gains remain small because many content changes appear only in states that short recordings of a fresh game rarely reach; target-specific probes or state injection that reach these states are a natural next step.

{
    \small
    \bibliographystyle{playcompass}
    \bibliography{main}
}

\end{document}